\documentclass{article}

\usepackage{arxiv}

\usepackage[utf8]{inputenc} %
\usepackage[T1]{fontenc}    %
\usepackage{lineno}
\usepackage[bookmarks=false]{hyperref}
\usepackage{url}            %
\usepackage{booktabs}       %
\usepackage{amsmath,amsfonts,amssymb}     %
\usepackage{nicefrac}       %
\usepackage{microtype}      %
\usepackage{cleveref}       %
\usepackage{lipsum}         %
\usepackage{graphicx}
\usepackage{csquotes}
\usepackage{doi}
\usepackage{cleveref}       %
\usepackage{dirtytalk}
\hypersetup{colorlinks,linkcolor={blue},citecolor={blue},urlcolor={black}}  
\usepackage[english]{babel}
\usepackage[nottoc]{tocbibind}
\usepackage{caption}
\usepackage{xcolor}
\usepackage{mathtools}
\usepackage{cleveref}       %

\title{Manually Selecting The Data Function for Supervised Learning of small datasets}

\date{}

\author{ \href{https://orcid.org/0000-0002-6412-9554}{\includegraphics[scale=0.06]{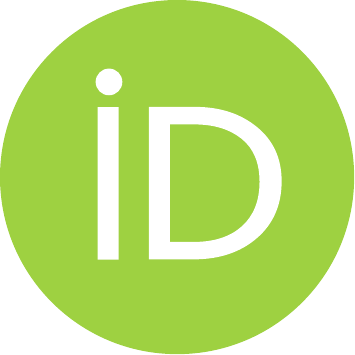}\hspace{1mm}Amir Khanjari}\\
	Department of Statistics\\
	Shahid Beheshti University\\
	Tehran, Iran \\
	\texttt{amir.khanjari.stat@gmail.com} \\
	\And
	Saeid Pourmand \\
	Department of Statistics\\
	Shahid Beheshti University\\
	Tehran, Iran \\
	\texttt{saeidpourmand1@gmail.com} \\
	\And
	\href{https://orcid.org/0000-0001-6917-1885}{\includegraphics[scale=0.06]{orcid.pdf}\hspace{1mm}Mohammad Reza Faridrohani} \\
	Department of Statistics\\
	Shahid Beheshti University\\
	Tehran, Iran \\
	\texttt{m\_faridrohani@sbu.ac.ir
	} \\
}
\renewcommand{\headeright}{}
\renewcommand{\undertitle}{}
\renewcommand{\shorttitle}

\begin{document}
	\maketitle
	
	\begin{abstract}
		Supervised learning problems may become ill-posed when there is a lack of information, resulting in unstable and non-unique solutions. However, instead of solely relying on regularization, initializing an informative ill-posed operator is akin to posing better questions to achieve more accurate answers. The Fredholm integral equation of the first kind (FIFK) is a reliable ill-posed operator that can integrate distributions and prior knowledge as input information. By incorporating input distributions and prior knowledge, the FIFK operator can address the limitations of using high-dimensional input distributions by semi-supervised assumptions, leading to more precise approximations of the integral operator. Additionally, the FIFK's incorporation of probabilistic principles can further enhance the accuracy and effectiveness of solutions. In cases of noisy operator equations and limited data, the FIFK's flexibility in defining problems using prior information or cross-validation with various kernel designs is especially advantageous. This capability allows for detailed problem definitions and facilitates achieving high levels of accuracy and stability in solutions.\\
		In our study, we examined the FIFK through two different approaches. Firstly, we implemented a semi-supervised assumption by using the same Fredholm operator kernel and data function kernel and incorporating unlabeled information. Secondly, we used the MSDF method, which involves selecting different kernels on both sides of the equation to define when the mapping kernel is different from the data function kernel. To assess the effectiveness of the FIFK and the proposed methods in solving ill-posed problems, we conducted experiments on a real-world dataset. Our goal was to compare the performance of these methods against the widely used least-squares method and other comparable methods.
	\end{abstract}


	\section{Introduction}
	The idea of supervised machine learning is based on the determination of a dependency function by a risk functional between two (or multiple) features. From a statistical inference perspective, a consistence approximation of genuine risk functional is carried out if the underlying data-generating process (DGP) can be estimated empirically with sufficient information or, specifically, if the nominated set of functions has a finite VC dimension. When there is ample data, consistency in the inductive principle can be achieved independently of the DGP.\\
	Collecting sufficient amounts of training data in order to make a model might be costly or time-consuming. For instance, in material science \cite{small, small2}, neuroimaging \cite{nero}, biomedical engineering \cite{Bio}, and many other subjects, it is pretty likely that not enough data exists. As a result, the efficacy of machine learning foundations is endangered, and their potent theorems cannot be successfully triggered. Therefore, insufficient information in supervised machine learning reflects a real challenge in constructing a good model for generalization using general principles. Notwithstanding the issue, we should still be able to estimate the desired function with just a minimal amount of data about the DGP.
	From the idea of Vapnik \cite{Vmain} for a small dataset in machine learning:
	\begin{displayquote}
		\emph{if you possess a restricted amount of information for solving some problem, try to solve the problem directly and never solve a more general problem as an intermediate step. The available information may be sufficient for a direct solution but is insufficient for solving a more general intermediate problem.}
	\end{displayquote}
	In other words, when working with a limited dataset, it is important to solve the problem of interest through a collection of related, straightforward statistical frameworks. Indeed, an approach that allows to select a framework that is specifically tailored to the problem at hand and can be modified as desired. Hence, in supervised machine learning, direct estimation of conditional density, conditional probability (classification), or regression problem\footnote{Regression function is one of conditional denisty characteristics. Despite the fact that we use regression function to avoid difficulties of conditional density estimation. However, the regression function can be formulated by its definition.} via integral operators convey possibilities in optional kernel design that can consider this matter \cite{Vnature}. However, these procedures are more complex than inductive principles and make us face ill-posed problems \cite{Vmain}. There are three requirements for inverse problems that must be fulfilled in order for them to be considered well-posed: existence, uniqueness, and stability of solutions. Suppose a solution exists, but we earnestly pursue the uniqueness and stability of the problem.
	The identifiability and estimability are the two questions that should be answered for uniqueness and stability, respectively \cite{What}. The identifiability is formalized as the injectivity of a specific forward mapping from the dependency to the datum. The estimability should be characterized as the continuity of the inverse to this mapping. The identification of machine learning is reflected in a relationship between machine learning and statistical inference in order to take advantage for finding an unique solution. However, the estimability of an inverse problem is the hardest part of this requirement and must be dealt with the instability of the estimator. On the other hand, the Tikhonov regularization that is addressed to solve all inverse ill-posed problems is pretty misleading. Indeed, before the regularization, the inverse operator should be defined and customized properly with integral equations. In other words, the inclination to obtain a high-level stability in inverse problems, a well-defined arbitrary framework for a specific problem should be followed before the regularization \cite{What}.\\
	However, there is an intimate relationship between the theory of inverse ill-posed and statistical learning problems. The learning theory from the identification point of view, is intrinsically probabilistic, whereas the inverse problem theory is primarily deterministic. The deterministic ill-posed problem on desired forward mapping is indeed noise-level/design fixed where the stochastic one in nature is determined by different levels of randomness \cite{Inverse}. This viewpoint, therefore, provides a route from noisy inputs to stochastic dependencies.\\
	The concept of the Fredholm integral of the first kind (FIFK) is central to the theory of inverse problems, and it is closely related to the theory of kernel-based learning. The Fredholm integral equations are a class of integral equations that are used to describe a wide range of inverse problems, including image restoration \cite{Image}, signal processing \cite{Signal}, and machine learning \cite{f1, f2}. One of the key advantages of using Fredholm integral equations to solve inverse problems is that they offer a high degree of flexibility in terms of the choice of kernel functions. This flexibility allows for the design of custom kernels that can take into account the specific features of the problem at hand, such as the noise level or the design of the experiment. Additionally, FIFK provides a natural framework for incorporating prior information about the unknown function, which can further improve the stability and accuracy of the solution. Solving Fredholm integral equations is not without its challenges. In particular, the inverse problem is often ill-posed, meaning that the solution is not unique or stable. To overcome this issue, various regularization techniques have been developed, such as Tikhonov regularization \cite{Oxf} and total variation regularization \cite{Vari}. These techniques aim to stabilize the inverse problem by introducing additional constraints on the solution.\\
	The Fredholm integral operator is a distribution-dependent operator that can be used to identify a desired function in the context of inverse ill-posed problems. Vapnik and Izmailov have demonstrated that by solving this operator, we can obtain the $\mathcal V$-Matrix property, which can be used to gain insight into the solution of ill-posed supervised machine learning problems \cite{VapnikV, Vmain, VS}. Specifically, the $\mathcal V$-Matrix method offers a clear understanding of the distributional location of each training sample, as it is represented by the coefficients of the loss function.\\
	However, the challenge that arises when working with a small dataset is that, unlike Empirical Risk Minimization (ERM) which is also take place for finite sample sizes, solving an ill-posed problem requires the infinite data limit of $n \rightarrow \infty$\footnote{There is no theory for solving an ill-posed problem with the finite sample size \cite{Vmain}} as a standard identifiability analysis \cite{What}. This requirement is also a prerequisite for estimability, making the problem even more difficult. Furthermore, in the predicting of effect from cause,  by independence of the mechanism, the input distribution contains no information about conditional probability \cite{Bern1}, meaning that the position of observations in the distribution does not affect the estimation of the solution. This can be particularly challenging when working with a distribution-dependent Fredholm operator. In this paper, we aim to apply semi-supervised algorithms to mitigate these issues as much as possible.\\
	We investigated two approaches for estimating conditional probability functions based on Fredholm integral equations.
	The first method is inspired by the $\mathcal{V}$-Matrix method \cite{VapnikV} but we have developed it into a semi-supervised version based on domain adaptation, so-called the S$\mathcal V$-Matrix method. This approach utilizes the same kernels as the Fredholm integral equation, including the operator kernel and the kernel associated with the data distribution. By assuming a semi-supervised setup, we aim to incorporate distribution properties into the problem and demonstrate the effectiveness of this method. Secondly, the main focus of this paper is to propose a solution for approximating the Fredholm Integral Equation of the First Kind (FIFK) using a combination of labeled and unlabeled observations. By using different kernels for the operator and data function, we aim to incorporate arbitrary statistical inferences for data distribution in machine learning. Additionally, by designing arbitrary data distributions using kernels, we aim to achieve stability in inverse problems, even when only a small amount of labeled data is available.\\
	The primary focus of this study is to establish a connection between inverse ill-posed problems, statistical inference, supervised, and semi-supervised learning. The goal is to estimate the desired solution using probability distributions and non-parametric statistics when labeled data is limited. This approach can only be successful if the estimability and identifiability of the ill-posed problem are ensured.\\
	In the next section of this paper, we will review existing literature on semi-supervised learning and its underlying assumptions. We will examine the connection between semi-supervised learning and ill-posed integral operators and how it can aid in solving these operators with high levels of convergence. Our methodology will involve developing the concept of conditional probability estimation using semi-supervised learning. This study will delve into the technical aspects of these ideas and provide references for further theoretical context. Furthermore, we will conduct experimental validation using real-world datasets for binary classification accuracy. The accuracy of eight different methods will be tested using datasets with a small proportion of training data. Additionally, we will investigate the impact of using unlabeled observations on the accuracy of the methods. For this purpose, five datasets will be selected for five semi-supervised methods and the classification accuracy will be evaluated in relation to the number of unlabeled observations used.

	\section{Related works in semi-supervised learning on integral equations}\label{related}
	Semi-supervised learning (SSL) describes a setting where in addition to a labeled data
	${\cal D}_{\ell} = \lbrace x_i,y_i \rbrace _{i = 1}^{n_\ell} $,
	one can obtain or already has access to rich information of unlabeled data
	${\cal D}_{u} = \lbrace x_i \rbrace _{i = n_\ell+1}^{n_{\ell}+n_u} $
	sampled from a similar distribution
	$P(x)$. The aim of SSL is the same as supervised learning: estimating the desired function
	$f$
	from points in a set
	$X$. Nevertheless, in SSL, one wants to make a better prediction from additional unlabeled information. One of the very first works on SSL was transductive learning, where the idea was to estimate a suitable function just for an unlabeled set rather than making a good universal prediction in induction principle \cite{Vapnik1}. However, in recent years, a comprehensive survey on semi-supervised learning was published [see Chapelle e al., 2006 and Zhu e al., 2002] \cite{Chap, Zhu2}, in which they provided clear study bases to explore SSL in transductive and inductive principles. After Zhu's survey analysis (2008) \cite{Zhu}, some great developments in SSL approaches have been proposed, and existing methods have been developed theoretically: Including EM with generative mixture models \cite{Fuji}, self-training \cite{Yar, Ril}, co-training \cite{Mich}, transductive SVMs \cite{Vmain}, and graph-based techniques \cite{GB, Belkin1, Belkin2, Par}.\\
	Nevertheless, there are controversial conflicts when the SSL approaches can make a better generalization despite the independence between the input distribution
	$P(x)$
	and conditional distribution
	$P(y|x)$ (or it's expectation) \cite{Bern1}. Hence, there is a non-negligible condition that states that the underlying data distribution
	$P(x)$
	must contain information about the conditional probability function \cite{Van}.
	This condition motivates the fattening of the data distribution toward unlabeled information through the semi-supervised learning method. Otherwise, we only pay the redundant and costly computation expenses without any gain, and in some cases, it is even destructive \cite{Zhu, Chap}. Therefore, in semi-supervised learning literature, the expected interaction between
	$P(x)$
	and
	$P(y|x)$
	can be configured using three important assumptions; the smoothness assumption, the low-density assumption and the manifold assumption. Lafferty and Wasserman \cite{Wasserman} define:
	the desired function
	$f$ (i.e, $P(y|x)$)
	is very smooth where the data function
	$P(x)$
	is large.
	In other words, if the connection path (for example, discrepancy function) between two observations
	$x$
	and
	$x'$
	on which
	$P(x)$
	is large, then the
	$y$
	and
	$y'$
	should lie within the same cluster.\\
	Basis on the smoothness definition, the problem should be equipped with available information in contrast to supervised learning methods that ignore the input distribution. Additionally, the low-density assumption definition pursues the same structure on input data function in which the decision boundary should lie in the low-density regions \cite{Chap2, Van}.\\
	However, a very important assumption that is important in this article is the manifold assumption: The high-dimensional data distribution function
	$P(x)$
	resides on a low-dimensional manifold \cite{Urak, Belkin2}.
	The above definition is an excellent motivation for semi-supervised learning using labeled and unlabeled information to enforce the function
	$f$
	on data manifolds. This flow can be characterized as the integral operator with respect to underlying data distribution. Let us define a distribution-dependent functional that preserves the geometry of probability distribution. Consider the distribution-dependent functional
	\begin{equation}
		R_P (f) = \int_{x \in {\mathbb{M}}} f(x) \Delta_{\mathbb{M}} f(x) dP(x) = \int_{x \in {\mathbb{M}}} f(x) ||\nabla_{\mathbb{M}}^2 f|| dP(x),
	\end{equation}
	where
	$\Delta_{\mathbb{M}}$
	is the  Laplace$-$Beltrami operator on the manifold
	${\mathbb{M}} \in \cal X$
	with respect to the probability distribution
	$P$
	and
	$\Delta_{\mathbb{M}} f=-div(\nabla_{\mathbb{M}} f)$. Above integral indicates
	the gradient $\nabla_{\mathbb{M}} f(x)$
	should be smooth where the probability distribution
	$P$
	at point
	$x$
	is large \cite{Par}.\\
	However, Belkin et al. (2006) \cite{Belkin1} introduced the Laplace-Beltrami operator as an additional regularization functional to exploit the geometry of marginal distribution. They showed controlling the complexity of the desired function in the intrinsic geometry of
	$P$
	on the kernel-based frameworks, improves the performance of the conditional probability estimation (or its family) under semi-supervised learning assumptions. Hence, based on labeled and unlabeled observations, the approximated Laplace-Beltrami operator embeds the information of underlying high-dimensional distribution into the coefficient matrix as a regularization term \cite{Belkin2}. The framework of manifold regularization can be formulated as follows,
	\begin{equation}
		\underset{f \in \cal M}{min} \sum_{i=1}^{n_{\ell}} L(y_i, f(x_i)) + \frac{C_1}{(n_{\ell} + n_u)} ||f||^2_{I} + C_2 ||f||^2_{T},
	\end{equation}
	where
	\begin{equation}
		||f||^2_{I} = \sum_{i=1}^{n} \sum_{j=1}^{n} w_{ij} (f(x_i)-f(x_j))^2, \hspace{0.5cm} n = n_{\ell} + n_u.
	\end{equation}
	The function
	$L$
	can be any desired loss function, such as the hinge loss
	$ max \{ 0,1-y_i f(x_i) \} $
	for the support vector machine or the least square method
	$(y_i-f(x_i))^2$; $C_1, C_2 > 0 $ are the tuning parameters corresponding to manifold regularization
	$||f||^2_{I}$
	and Tikhonov regularization functional
	$||f||^2_{T}$, respectively. Here, the admissible set of functions
	$\{f(x) \}$
	is defined in rich reproducing kernel Hilbert space (RKHS) to follow Representer theorem.\\
	However, instead of the Laplace$-$Beltrami operator, we can consider another data-depend operator so-called ill-posed Fredholm integral operator \cite{f3}. The Fredholm integral of the first kind (FIFK) is defined as:
	\begin{equation}\label{FF}
		{\cal K}_P f = \int_{\mathbb{M}} K_{\mathbf{F}}(T,x) f(x) dP(x) = G(T),
	\end{equation}
	Where
	$K(t,x)$
	is the semi-definite function is called the Fredholm integral kernel, and in this paper, we assume the function
	$f(x) = p(y|x)$
	is the conditional probability for classification task. Here,
	${\cal K}_P$
	is the operator that maps the set of functions
	$\{f(x)\}$
	into the set of data function
	$\{G(T) \}$
	associated with kernel
	$K(T,x)$. By the law of large numbers for approximation of eq(\ref{FF}) with respect to the probability distribution
	$P(x)$ based on labeled and unlabeled observations can be obtained as
	\begin{equation}
		{\cal K}_{\hat{P}_{n}} f = \frac{1}{n} \sum_{j=1}^{n} K_{\mathbf{F}}(T,x_j) f(x_j), \hspace{4mm} n_\ell + n_u=n.
	\end{equation}
	The above approximation provides an elegant way (different from the Laplace-Balterami operator) of merging unlabeled information into the learning machine. In this Fredholm framework, the Fredholm space function
	${\cal N}_F = \{{\cal K}_P f;  f \in \mathcal H  \}$, where
	$\mathcal H$
	is a
	$\cal {RKHS}$
	as classification and regression functions. Noting that, in a different form, the function space $\cal {RKHS}$, the Fredholm function space is density dependent. The right hand-side of the equation, $G(T)$ represents the data function in inverse problem. Regardless of the stochastic form of the data function, instead the fixed and noise-free responses $Y$ are selected ($y \approx G(T)$) as a exact datum and image of Fredholm operator solution.\\
	The following optimization in the Fredholm learning framework is similar to supervised machine learning risk functionals:
	\begin{equation}\label{oldF}
		\hat{f} = arg \hspace{1mm} \underset{f \in \cal H}{min} \frac{1}{n_\ell} \sum_{i=1}^{n_\ell} \big( {\cal K}_{{P}_{x_\ell}} f(x_i) - y_i  \big)^2 + \lambda ||f||_T^2.
	\end{equation}
	From the using representer theorem, the desired solution in
	$\cal {RKHS}$
	as follows:
	\[
	f(x) = \frac{1}{n} \sum_{i=1}^{n} \alpha_i K (x,x_i),
	\]
	the parameters
	$\boldsymbol{\alpha}$ has closed-form solution:
	\[
	\boldsymbol{\alpha} = (K^T_n K_n  K_{\cal H} + \lambda I)^{-1} K^T_n \boldsymbol{y},
	\]
	where $[K_{\mathbf{F}}]_{ij} = K(t_i,x_j)$ for
	$1 \leq i \leq n_\ell$,
	$1 \leq j \leq n $, and $[K]_{ij} = K_{\cal H} (x_i,x_j)$
	for $1 \leq i,j \leq n $. The Fredholm learning is studied under the noise assumption \cite{f2} in contrast to cluster assumption or manifold assumption in semi-supervised learning: The low variance directions within every sample neighbor are unimportant in terms of class labels and can be regarded as noise. Based on this assumption, the Fredholm integral has noise-suppression capacity. Particularly, when samples are contaminated by noise, it will give a more accurate estimate of the original data, and the resulting desired function will be more closely related to the real form \cite{f1}.\\
	However, in learning from examples based on statistical inverse problem with respect to the probabilistic perspective, our goal is to find a significant approximation to the solution of the inverse problem where the right-hand side is defined stochastically by indirect measurement $G(T)$, rather than finding a stable approximation to the solution of the discrete Problem (\ref{oldF}) \cite{Inverse}. The concept of stochastic ill-posed problem arises when both the operator and the data function $G(T)$ are characterized stochastically \cite{Vmain}. Solving this type of problem requires obtaining reasonable approximations for both the operator and the data function. The Fredholm integral operator is utilized in the operator equation with the aim of the identification solution in machine learning, using the direct definition of conditional probability (in the next section we define the problem in details). However, Vapnik and Izmailov \cite{VapnikV} proposed that the data function $G(T)$ can be modeled as a cumulative distribution function, which benefits from the empirical inference with a high rate of convergence, as per the Glivenko–Cantelli theorem. Using this theory, a coefficient known as the $\mathcal{V}$-Matrix has been derived from the Fredholm integral equation, which represents the relationships between the observations (mutual positions of observations) via indicator functions.\\
	$\mathcal{V}$-Matrix is a $n_\ell \times n_\ell$ matrix that can be defined in two main forms in a d-dimensional case:
	\begin{itemize}
		\item {\bf Uniform indicator $\mathcal{V}$-Matrix:}
		\begin{flalign}\label{uiv}
			&\mathcal{V}_{ij} = \prod_{k=1}^{d} (C^k - max \{ x^k_{i},x^k_{j} \}),\hspace{3mm} C^k = Max \{x^k \},\hspace{3mm} i,j=1,...,n_\ell.&
		\end{flalign}
		Remark. Instead of $\prod$ in eq(\ref{uiv}), one can replace $\sum$ to obtain additive $\mathcal{V}$-Matrix for computational reasons.
	\end{itemize}
	The $\mathcal{V}$-Matrix can be obtained with a CDF metric on the residual of the operator equation:  
	\begin{itemize}
		\item {\bf CDF-measure Indicator $\mathcal{V}$-Matrix:}
		\begin{flalign}
			&\mathcal{V}_{ij} = \sum_{t=1}^{n_\ell} \prod_{k=1}^{d} I\big(x_t^k - max \{x^k_i,x^k_j \}\big) &
		\end{flalign}
		Where $I(z)=1$ for $z \geq 0$ is an indicator function.
	\end{itemize}
	These $\mathcal{V}$-Matrices are used in some works to re-weight least-square such as parametric regression estimation \cite{Vamir}, support vector regression \cite{V3}, SVM \cite{Vcf, VS, VapnikV} and gradient boosting \cite{Maz}. All methods are supervised machine learning, and they attempted to boost the accuracy of the model just based on training data.\\
	However, it is not clear why and when the distribution dependent Fredholm operator equation can help to boost the accuracy. As we mentioned, for incorporating the input probability distribution or in here specifically the location of observations in distribution, we need semi-supervised assumptions. In $\mathcal{V}$-Matrix method it seems we can use it in domain adaptation and robust it against covariate shift problems \cite{Maz}. 
	
	\section{Methodology}
	In statistical learning theory literature, the goal of machine learning is divided into two main ways: Imitation and identification of true supervisor's operator \cite{Vmain}, Where the supervisor phenomena is associated with the conditional probability $p(y|x)$. From an identification learning machine point of view, the goal is to define the supervisor directly and discover a method for achieving a close identifiable approximation. However, in the imitation perspective, the goal is just to define a set of rules for prediction, no matter what is the form of the supervisor. Let's consider the direct definition of conditional probability $p(y=1|x)p(x)=p(y=1,x)$ for classification, where $y \in \{0,1\}$ and $x \in X$. In order to an identifiable and estimable (high-level of stability) estimator for conditional probability, the first step is to define the appropriate forward mapping. Instead of functional eq(\ref{FF}), we define the operator equation
	\begin{equation}\label{ill}
		{\cal K} f = G,
	\end{equation}
	where
	${\cal K}$
	is an operator that maps a set of conditional probability distribution functions
	$\{f\}$
	in object space
	$\mathcal{H}$
	to a set of data functions
	$G$
	in image space
	$L^2(X,\nu)$. Let $L^2(X,\nu)$ be the Hilbert space of square-integrable functions on $X$ with respect to the
	marginal measure $\nu$.
	The associated kernel $K$ in this mapping is positive definite kernel (not necessarily symmetric). The fact that if $K$ is continuous then ensure the injectivity of $\mathcal K$ and if non-negative measure $\nu$ then is not degenerated, Therefore, the inverse of above forward mapping set out identified solution $f$ \cite{Inverse}.\\
	One of the prominent method for defining well-defined operator for eq (\ref{ill}), is the Fredholm integral equation of the first kind. In deterministic Fredholm learning methods \cite{f2}, the operator equation ${\mathcal K}_P f \approx y$ is used to capture the noise in the observations by using a kernel $K$ associated with the marginal distribution $P$. However, this approach does not fully capture the complexity of the inverse problem in statistical learning problems, which are ill-posed problems with variable or error-prone data in both the operator $\mathcal K$ and the right-hand side data function. The traditional approach of approximating $G \approx y$ is not sufficient and a more robust method is needed to accurately solve the problem without just relying on the regularization.\\
	However, in this paper, the Fredholm operator that is dependent on the input distribution is defined \cite{VapnikV}, and its vigor pertains to how this function is approximated. On the other hand, this operator is intended to be extremely sensitive to even the most minor changes on the other side of the equation \cite{Fhand, Fhand2}.\\
	Now, in order to construct conditional probability as a Fredholm integral solution, according to the Bayes theorem the direct definition of the conditional probability function is defined as
	\begin{equation}
		p(y=1|x)p(x) = p(x,y=1).
	\end{equation}
	Suppose $K(T,x)$ is an arbitrary semidefinite kernel function for all $x$ and $T$ in $X$, therefore
	\begin{equation}\label{Fmain}
		\int_{\mathcal X}	K(T,x)p(y=1|x)p(x)dx =\int_{\mathcal X} K(T,x)p(x,y=1)dx.
	\end{equation}
	Let denote $X$ a subset of high d-dimensional Euclidean space $\mathbb{R}^d$. The variable T represents an extra layer of information that is captured by the kernel function. The kernel function $K(T,x)$, maps both the inputs $x$ and $T$ into a high-dimensional feature space, where the integral operator acts on the function $p(y=1|x)$ to produce the output $G(T)$. It allows for the capture of variations and noise as a weighting function in the data that may not be captured by the input $x$ alone. By rewriting the above equation, we have
	\begin{equation}\label{mainf}
		\int_{\mathcal X} K(T,x) f(x) dP(x) =G(T),
	\end{equation}
	where
	$f(x) = p(y=1|x)$
	is the conditional probability in Hilbert space $\cal H$ for binary classification as a solution of integral integral operator with a kernel $K$ is denoted by $\mathcal{K}_P$ is associated with probability measure (marginal distribution) $P$. The other hand-side of the above equation, the function
	\begin{equation}
		G(T) = E[K(T,X=x) | Y=1] = \int_{\mathcal{X}} K(T,x) p(y=1,x)dx,
	\end{equation}
	Recall that T is a random variable that makes $G(T)$ is a stochastic regression function\footnote{The $G(T)$ is a function that maps $T$ to a predicted value for the kernel function $K(T,x)$ at that point $T$. This predicted value is not a fixed constant, but rather a random variable that follows some high-dimensional probability distribution. Therefore, $G(T)$ can be thought of as a random design regression function.}. The noise and uncertainty in the problem can be controlled by selecting a good form of $G(T)$ as a regression function because it provides information about the expected behavior of $f(x)$ based on the prior knowledge of the problem. Therefore, in contrast to the deterministic Fredholm integral equation property\footnote{The right-hand side of the equation is known exactly and is not a random variable, the solution is simply the expected value of $f(x)$ given the observed data, which is represented by $y$ in the equation $E[f(x)|y]$.}, the prior knowledge is encoded in $G(T)$ and the expected behavior of $f(x)$ is given by $E[f(x)|G(T)]$. Additionally, this is a property of the Fredholm integral equation solution because it provides a probabilistic interpretation of the solution, which can be useful for understanding the noise and uncertainty of the problem.\\
	In the other hand, $G$ represents an integral operator that maps the joint distribution to the data $G$ with kernel $K(.,x)$. Selecting an appropriate kernel function is indeed an important step in Bayesian approach to machine learning \cite{Bern2}. The kernel function encodes the prior knowledge or assumptions about the form of the underlying relationship between the input and output variables. By integrating the joint distribution $p(x,y=1)$ with the kernel function, the Bayesian approach models the data function $G$, which is a probabilistic representation of the observed data.

	In order to find the Fredholm integral equation solution, implementation of least-square method guarantees the uniqueness and existence of the solution:
	\begin{equation}
		\underset{f \in \mathcal H}{min} ||\mathcal{K}_P f(T)-G(T)||^2_{L^2(X,\nu)}.
	\end{equation}
	The function $G$ can be consider as a regularizer for Fredholm integral equation. With selecting a decent function for G, estimation of conditional probability is extremely dependent on data function. In other words, a small changes of data function can provide big changes in solution and this is a big challenge in estimability. We need to guarantee the stability of solution with respect to different choices of G. Then, we can check the expected behavior of solution respect to the data function. Indeed, if the function $G$ is selected properly and has useful prior information, then we can reduce degree of the enforcement of the regularization term.\\
	Therefore, in order to approximate the integral eq(\ref{mainf}), we need to implement Tikhonov regularization (TR) to consider the stability of the solution. TR transforms the Fredholm integral equation of the first kind to the second one \cite{Fhand3}, which enables us to apply the discrete counterpart of the integral operator \cite{f2} and use the approximated cumulative distribution function for the right-hand side. Then, the TR functional is defined as
	\begin{equation}\label{op}
		R(f) = \rho^2_{{L^2(X,\nu)}} ({\mathcal{K}}_{P_{\ell}} f,F_\ell) + \lambda W_{{L^2(X,\nu)}}(f),
	\end{equation}
	where
	$W_{{L^2(X,\nu)}}(f)$
	is the regularization functional in the Hilbert space
	${L^2(X,\nu)}$
	with associated regularizer parameter
	$\lambda > 0$
	and
	$\rho^2_{{L^2(X,\nu)}}$
	is the discrepancy functional is defined in the space
	${L^2(X,\nu)}$.\\
	
	In this paper, we peruse two different approaches for approximation of Fredholm integral equation and conditional probability estimation. 
	\begin{enumerate}
		\item {\bf Semi-Supervised $\mathcal{V}$-Matrix(SV-M):}
		Approximation of the integral operator and the right hand-side using only labeled observations when the kernel functions in both of sides of the equation are same. However, we will implement the unlabeled observations for measure of discrepancy of the equation based on extra layer $T$ that is captured by $K(T,.)$, to follow the semi-supervised learning.
		
		\item {\bf MSDF (Manually selecting data function):}
		Approximation of the integral operator by Using both unlabeled and labeled data points for capturing semi-supervised assumption where the kernels in Fredholm equation is not the same. The data function kernel is selected based on different interpretation from the operator kernel.
	\end{enumerate}
	We start with case(1) which is inspired by the works of Vapnik and Izmailov \cite{VapnikV} introduced $\mathcal{V}$-Matrix method for solving ill-posed statistical learning problems.
	
	\subsection{The $\mathcal{V}$-Matrix Method}\label{sec3.1}
	Here, we consider d-dimensional labeled observations $\mathbf{x}_{n_\ell}=(x_1,x_2,\dots,x_{n_\ell})$ to approximate the operator equation ${\cal K}_P f=G$. Indeed, the noisy operator $\mathcal{K}_{\delta}$ that is approximated is defined as
	\begin{equation}
		\mathcal{K}_{\delta}: \mathcal{H} \rightarrow \mathcal{M},
	\end{equation}
	where $\mathcal{M}$ is the noisy data space and the noisy data function $G_{\delta}$ belongs to this noisy space.
	In other words, by the law of large numbers, by using labeled data, the approximation of the eq(\ref{mainf}) is defined as
	\begin{equation}\label{AF}
		\sum_{i=1}^{n_\ell} K(T,x_i)f(x_i) \approx \sum_{j=1}^{n_\ell} y_j K(T,x_j).
	\end{equation}
	The optimization problem (\ref{op}) based on the above approximation depends on the random variable $s \in \cal X$ by associated with any desired kernel $K(x_i,s)$ as $x_i \in \mathcal{D}_{\ell}$.
	First, the noisy data space $\mathcal{M}$ is different from the exact data space $L^2(\cal X,\nu)$ so that $\cal K$ and $\mathcal{K}_x$ belong to different spaces. However, in contrast to finite-dimensional noisy space, we consider high-dimensional $L^2(\cal X,\nu)$ space for discrepancy of eq(\ref{AF}) in order to descent estimate of an identified solution.
	Therefore, the distance measure $\rho^2(.,.)$ on the space ${L^2(\cal X,\nu)}$ is defined as
	\begin{equation}\label{rho}
		\rho^2_{L^2(\cal X,\nu)} = \int_{\cal X} \Big({\mathcal{K}}_{P_{x_\ell}} f(T) - {G_x(T)} \Big)^2 d \nu(T).
	\end{equation}
	Hence after some computations, the new risk functional is defined as
	\begin{equation}\label{vrisk}
		R(f) = \sum_{i,j = 1}^{n_\ell} \Big(y_i - f(x_i) \Big)\Big(y_j - f(x_j) \Big)\mathcal{V}_{ij} + \lambda W_{L^2(\cal X,\nu)}(f),
	\end{equation}
	where the $\lambda > 0$ is the tuning parameter and $W$ is the regularization functional on the space $L^2(\cal X,\nu)$. The $\mathcal{V}_{ij}$
	is the mutual kernel matrix inspired by Vapnik and Izamailov \cite{VapnikV} as a $\cal{V}$-Matrix (Vapnik-Matrix) concept and is defined as
	\begin{equation}\label{TF}
		\mathcal{V}_{ij} = \int_{\mathcal{X}} K(T,x_i)K(T,x_j) d \mu(T) = \langle K(T,x_i), K(T,x_j) \rangle_{L^2(\cal X,\nu)}.
	\end{equation}
	The $\mathcal{V}$-matrix is an inner product\footnote{$\mathcal{V}_{ij}$ is symmetric, positive definite and has linearity in it's kernels for all $i,j$.} that measures the similarity between all pairs of observations, where a high value indicates a high similarity between the two observations, and a low value indicates a low similarity. For example, for $i$ and $j$ if the value of the $\mathcal{V}_{ij}$ is zero, it means that the observations $x_i$ and $x_j$ are orthogonal, or dissimilar. Therefore, when $\mathcal{V}_{ij}$ is zero it means that the quadratic optimization for that specific pair of observations $x_i$ and $x_j$ will have no effect on the final solution. This is because the observations are dissimilar and do not contribute to the overall similarity of the observations. The variable T, is known as a reference point or anchor that is used as a point of comparison for measuring the similarity between the training observations.\\
	However, the risk functional without regularization term can be re-written as
	\begin{equation}\label{eq21}
		R(f) = \sum_{i=j=1}^{n} (y_i - f(x_i))^2\mathcal{V}_{ii} + \sum_{i \neq j}^{n} (y_i - f(x_i))(y_j - f(x_j))\mathcal{V}_{ij}.
	\end{equation}
	When $i=j$, the risk functional reduces to a weighted least-squares problem, with the weight given by $\mathcal{V}_{ii}$. However, in the second term, which involves the sum over $i \neq j$, the residuals are multiplied by the off-diagonal elements $\mathcal{V}_{ij}$ of the matrix $\mathcal{V}$. This term captures the correlations between the residuals at different points, and takes into account the relationships between the residuals at each pair of points. As a result, the risk functional in the this method provides a more general and flexible way of measuring the discrepancy between the model and the data, as it allows for different weights to be assigned to the residuals at different pairs of points.\\
	The mutual residual term in the $\mathcal{V}$-Matrix method can affect the problem in several ways:
	
	\begin{figure}
		\centering
		\begin{minipage}{0.5\textwidth}
			\centering
			\includegraphics[width=\linewidth]{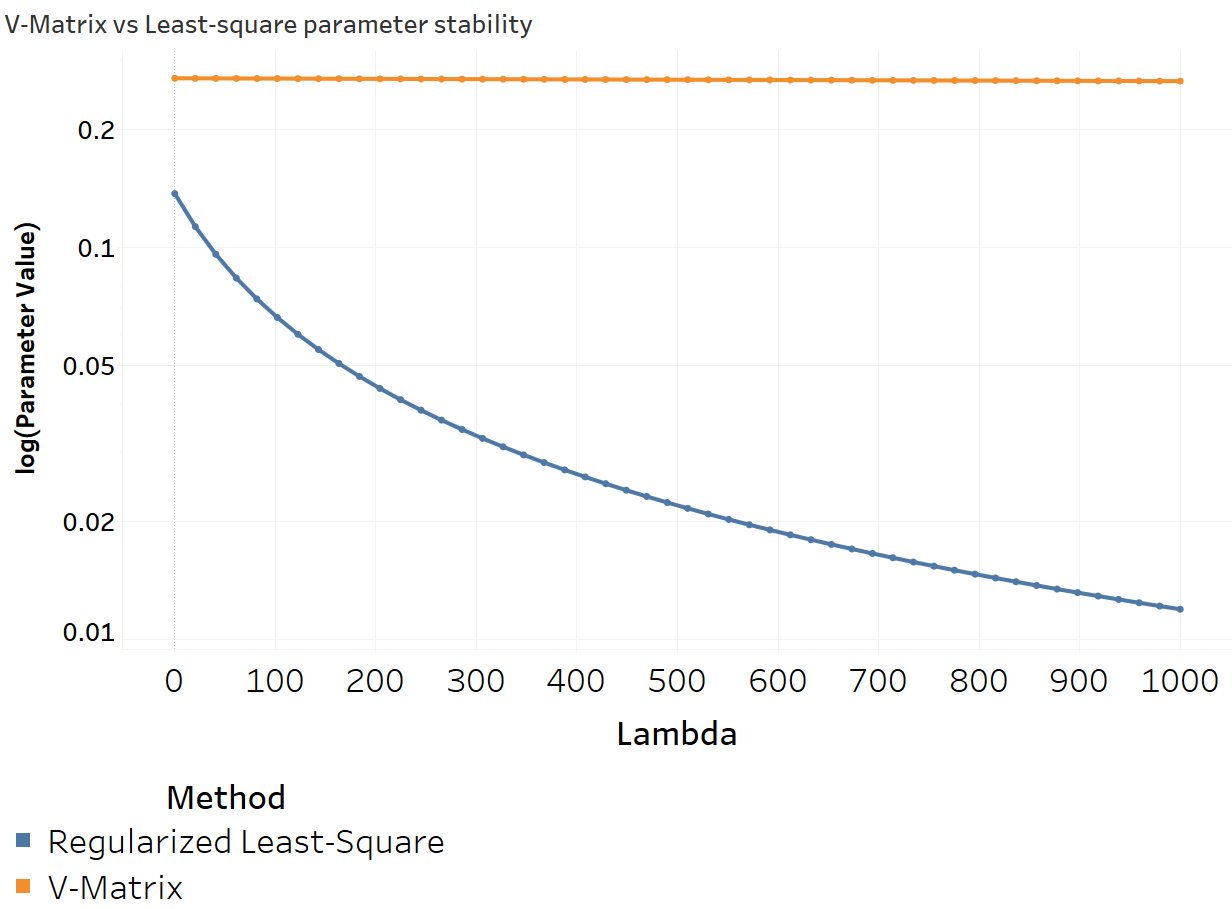}
		\end{minipage}%
		\begin{minipage}{0.5\textwidth}
			\centering
			\includegraphics[width=\linewidth]{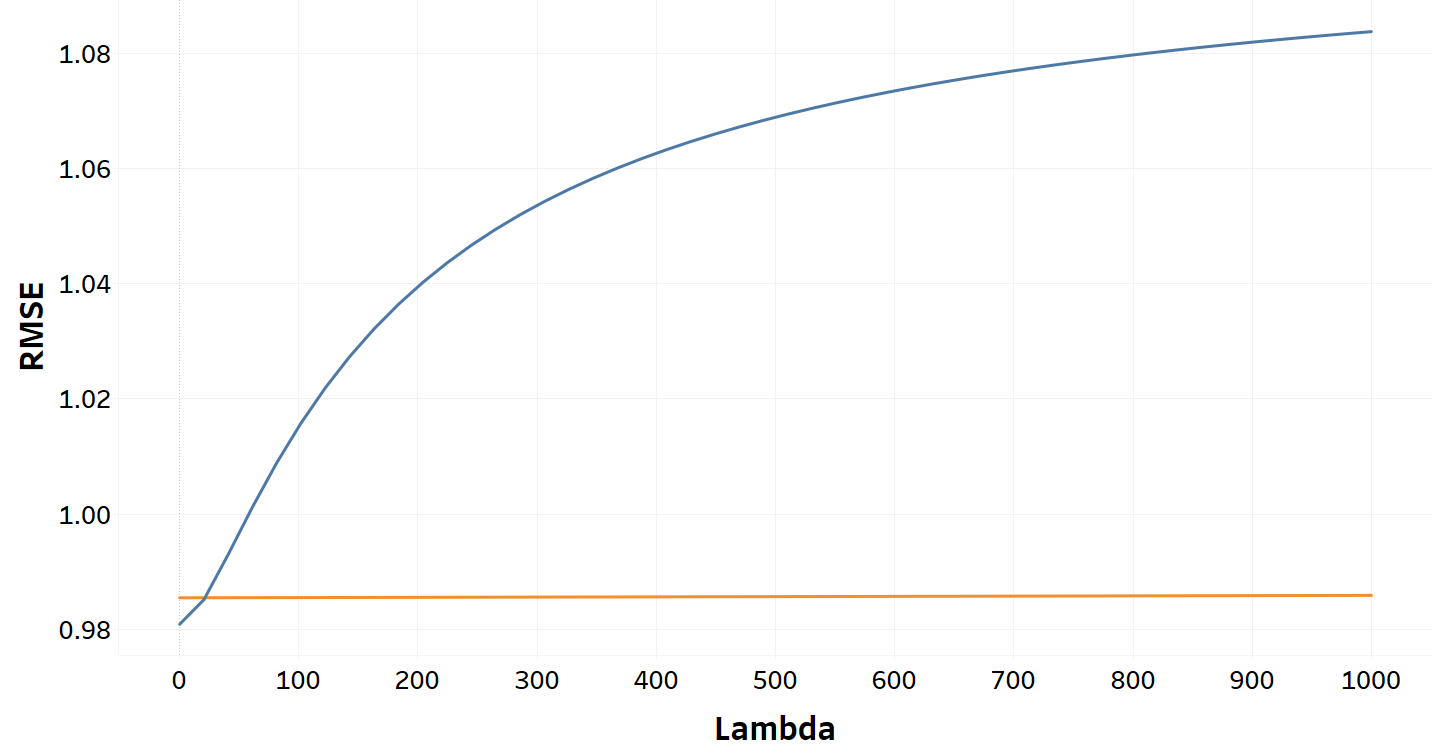}
		\end{minipage}
		\caption{The plot displays the stability of the parameter and accuracy estimated by $\mathcal{V}$-Matrix regression and regularized least-square for the linear model $y = x\beta + \epsilon$, where $x \sim N(0,1)$ and $\epsilon \sim N(0,1)$. The comparison is made by plotting the estimated parameter values obtained using the two methods.}
	\end{figure}

	\begin{itemize}
		\item \textbf{Robustness}: The mutual residual term in the risk functional takes into account the relationships between the residuals at different pairs of points, rather than just the individual residuals. This can lead to a more robust solution that is less sensitive to outliers or other types of errors in the data.
		
		\item \textbf{Model flexibility}: By including the mutual residual term, the $\mathcal{V}$-Matrix method provides a more general and flexible way of measuring the discrepancy between the model and the data. This allows for different weights to be assigned to the residuals at different pairs of points, which can result in a better fit to the data in cases where the residuals are not well-behaved or have a complex structure.

		\item \textbf{Solution quality}: The choice of kernel function and measure of distribution used to compute the matrix $\mathcal{V}$ can significantly affect the quality of the solution. The choice of these parameters determines the form of the mutual residual term and can influence the robustness, flexibility, and accuracy of the solution. It is important to choose these parameters carefully to ensure that the solution is well-behaved and provides a good fit to the data.
	\end{itemize}

	\subsubsection{$\cal{V}$-Matrix setting}
	Here the integral of the multiplication of two kernels
	$K(x_i,s)K(x_j,s) = K(s;x_i,x_j)$
	implies a unique kernel concerning the true measure $\mu$ (full probabilistic model with respect to a unique environment). However, to take first step for approximating of true $\cal{V}$, suppose the measure $\mu(s) = P(s)$ is a high-dimensional probability distribution that contains all information about the input distribution. The distribution $P(x)$ (distribution of training observations) is a subset of this high-dimensional probability. Therefore, the distribution of training observations might not contain all the information of the true distribution.\\
	Therefore, the new $\cal{V}$ is defined as a expectation of kernel
	$K(s;x_i,x_j)$
	w.r.t probability
	$P(T)$.
	Here, in order to approximate the probability
	$P(T)$
	based on Glivenko-Cantelli theorem where the points increase sufficiently (with labeled $\mathbf{x_{n_\ell}}$ and unlabeled $\mathbf{x_{n_u}}$) then the empirical estimation converges to the true $P(T)$. The approximated $\cal{V}$ is defined as
	\begin{equation}
		\hat{\mathcal{V}}_{ij} = \sum_{k=1}^{n}K(x_k;x_i,x_j), \hspace{5mm} i,j=1,...,n_{\ell}, \hspace{5mm} n_{\ell}+n_{u} =n
	\end{equation}
	This unmodified kernel should be customized to represent nodes (data points) and edges (similarities between points) by labeled and unlabeled observations. Here the convolution kernel 
	$K(x_k^h;x_i^h,x_j^h) = K(x_k^h- \{ x_i^h,x_j^h\} )$
	can be used in order to determine the distances between rich labeled-unlabeled nodes. Further, the cornerstone of this setting is in that: n linear combination of d-dimensional input space, which is captured by kernel functions are embedded in every indices
	$i,j$.\\
	This configuration is based on the covariate adaptation assumption in semi-supervised learning problems, which provides intuition about better estimation of the conditional probability for unseen observations. indeed, when we have limited observations, as the unseen observations might have a different distribution.
	\begin{figure}
		\begin{center}
			\includegraphics[width=0.8\linewidth]{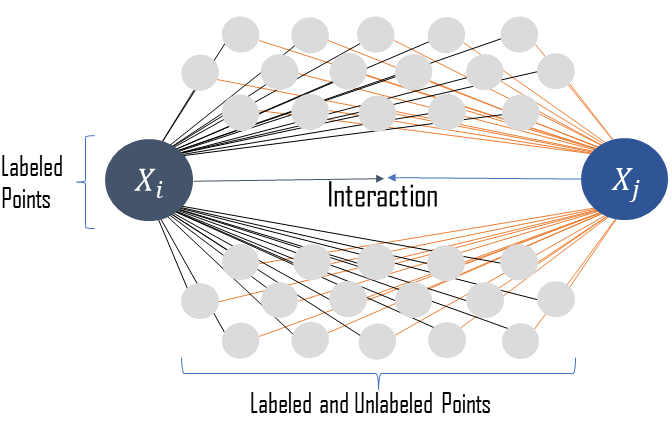}
			\caption{S$\mathcal{V}$-Matrix captures the geometric properties of labeled-unlabeled points and considers their interaction mutually for a high-dimensional linear combination of unlabeled points embedded in every labeled indices.}
		\end{center}	
	\end{figure}
	\begin{itemize}
		\item {\bf Semi-Supervised Indicator $\mathcal{V}$-Matrix(SI$\mathcal{V}$-M):}
		Let
		$K(t-x) = I(t-x)$
		is an indicator function for
		$t \ge x$. for this case,
		\begin{equation}\label{indicator1}
			\hat{\mathcal{V}}^{\cal{SIV}}_{ij} = \sum_{k=1}^{n} \prod_{h=1}^{d} I(x_k^h-x_i^h)I(x_k^h-x_j^h),
		\end{equation}
		Or
		\begin{equation}\label{indicator2}
			\hat{\mathcal{V}}^{\cal{SIV}}_{ij} = \sum_{k=1}^{n} \prod_{h=1}^{d} I \Big(x_k^h-Max \{x_i^h,x_j^h\}\Big)
		\end{equation}
	\end{itemize}
	SI$\mathcal{V}$-M compares the values of unlabeled observations in all dimensions with the values of labeled observations in all dimensions. Therefore, the indicator kernel is a way to measure the similarity between two observations ($x_i,x_j$) based on whether or not they have the same direction in all dimensions. In other words, the two observations have similar values for that feature, and the kernel is essentially counting how many features or dimensions the two observations have in common. Additionally, the selection of indicator functions for $\mathcal{V}$-Matrix kernels set out useful statistical properties results. If we select the indicator kernel function for $\mathcal{V}$-Matrix, the true SI$\mathcal{V}$-M is defined:
	\begin{equation}
		\mathcal{V}_{True}^{\cal{SIV}} (i,j) = \int I(T-x_i)I(T-x_j) d\mu(T),
	\end{equation}
	Therefore the following properties are valid\cite{Maz}.
	
	\begin{enumerate}
		\item $\mathcal{V}^{\cal{SIV}}$ is the minimum variance unbiased estimator (MVUE) of true $\mathcal{V}_{True}^{\cal{SIV}}$.
		\item According to the $\cal V$-risk functional(\ref{vrisk}) and previous result, The SI$\mathcal{V}$-M method strongly converges to the $\mathcal{V}_{True}^{\cal{SIV}}$ for enough large unlabeled points. Hence, there exist $\delta' > 0$
		\begin{equation}
			p(\big|\rho^2_{\mathcal{V}_{True}^{\cal{SIV}}} - \rho^2_{\hat{\mathcal{V}}^{\cal{SIV}}} \big| \leq \delta' \geq 1 - \frac{2}{n^2}
		\end{equation}
		Where $\delta' = \sqrt{\frac{log(n)}{n}}\sum_{i=1}^{n_\ell}\frac{|(y_i -f(x_i))(y_j-f(x_j))|}{n^2_\ell}$.
		\item Based on the Fredholm integral equation, the right hand-side of equation can be re-written as
		\begin{equation}
			G(T) = \int I(T-x) p(y=1,x)dx = \int_{x \le T} p(y=1,x)dx.
		\end{equation}
		Therefore the right hand-side of the Fredholm integral equation that defines our data function, is a cumulative distribution function $P(y=1,T \ge x)$.
		
		\item It can be used in ordinal data where the order of the observations is important and to compare the relative positions of observations even when the data is not numerical.
	\end{enumerate}

	\begin{itemize}
		\item {\bf Semi-supervised Gaussian $\mathcal{V}$-Matrix(SG$\mathcal{V}$-M):}	
		Let
		$K(t-x) = exp(-(t-x)^2/\sigma^2)$
		is a gaussian kernel where
		$\sigma^2 > 0$ is its the free parameter. Therefore,
		\begin{equation}
			\hat{\mathcal{V}}^{\cal{SGV}}_{ij} = \sum_{k=1}^{n}
			exp\Big(-\sum_{h=1}^{d}\Big[(x_k^h-x_i^h)^2+(x_k^h-x_j^h)^2 \Big]\frac{1}{\sigma^2} \Big)
		\end{equation}
	\end{itemize}
	The following results for SG$\mathcal{V}$-M for estimating the True $\mathcal{V}$-Matrix is valid.
	
	\begin{enumerate}
		\item The Gaussian kernel is infinitely differentiable, which means that the resulting SG$\mathcal{V}$-M will be smooth.
		\item The Gaussian kernel is a localized similarity measure, meaning that it gives more weight to unlabeled points that are close to labeled observation in feature space, and less weight to those that are farther apart. This results in a SG$\mathcal{V}$-M that is sensitive to local structure in the data.
		
		\item The Gaussian kernel is scale-invariant, which means that it is not sensitive to changes in the scale of the features. This results in a SG$\mathcal{V}$-M that is not sensitive to differences in the units of measurement of the features.
		\item The Gaussian kernel has a free parameter, $\sigma^2$, which controls the width of the kernel, with smaller values leading to a more localized similarity measure and larger values resulting in a more global similarity measure.
	\end{enumerate}

	\subsection{Conditional probability estimation with manually selecting the data function: MSDF method}
	In this section, a different strategy than the $\mathcal{V}$-Matrix method is sought to be implemented by altering the form of the data function in such a manner as to be equipped with a kernel that differs from that of the Fredholm operator kernel.\\
	Based on the section(\ref{sec3.1}), the noisy space of data function $\mathcal{M}$ is different form the exact data function space $L^2(X,\nu)$. Therefore the kernel in the function $G_{\delta}(T)$, can be any kernel function (in this setting, the kernel function does not have to be a positive semi-definite or even symmetric kernel) in Hilbert space $\mathcal{M}$. Therefore:
	\begin{equation}
		G_{\delta}(T) = \int_{X} K_{\mathbf{D}} (T,x) dP(y=1,x),
	\end{equation}
	where $K_{\mathbf{D}}(.,.)$ is the dedicated data function kernel. Here, instead of approximation (\ref{AF}), we implement labeled and unlabeled observations in order to descent approximation of integral \cite{f2}. Therefore
	\begin{equation}
		{\cal K}_{P_{n}}f=\frac{1}{n} \sum_{i=1}^{n} K_{\mathbf{F}}(T,x_i)f(x_i), \hspace{5mm} n_\ell + n_u = n.
	\end{equation}
	However, the right-hand side of the Fredholm equation is approximated by labeled data for following noise assumption in semi-supervised learning \cite{f2, f1, f3}. Then, in order to estimate function $f_{\mathcal{H}}$, the risk functional (\ref{op}) is defined as
	\begin{equation}
		\rho^2_{E_2}= \int \big( \frac{1}{n} \sum_{i=1}^{n} K_{\mathbf{F}}(T,x_i)f(x_i)- \frac{1}{n_\ell} \sum_{j=1}^{n_\ell}K_{\mathbf{D}}(T,x_j)y_j  \big)^2 dP(T)
	\end{equation}
	Here, the probability $P(T)$ is approximated by labeled observations on space $L^2(X,\nu)$. Hence
	\begin{equation}
		\begin{split}
			\rho^2_{L^2(X,\nu)} = \frac{1}{n^2} \sum_{t=1}^{n_\ell} \sum_{i=1}^{n} K_{\mathbf{F}} (x_t,x_i) K_{\mathbf{F}}(x_t,x_i) f(x_i)f(x_i) \\
			-2 \frac{1}{nn_\ell} \sum_{t=1}^{n_\ell} \sum_{i=1}^{n}\sum_{j=1}^{n_\ell} K_{\mathbf{F}} (x_t,x_i) K_{\mathbf{D}} (x_t,x_j) f(x_i)y_j \\
			+ \frac{1}{n_\ell^2} \sum_{t=1}^{n_\ell} \sum_{j=1}^{n_\ell} K_{\mathbf{D}}(x_t,x_j) K_{\mathbf{D}}(x_t,x_j) y_j y_j,
		\end{split}
	\end{equation} 
	where the minimizer of our optimization problem admits an expansion \cite{Back, f1}
	\begin{equation}\label{sol}
		f^{*}(.)=\sum_{i=1}^{n}\alpha_iK(.,x_i), \hspace{5mm} n_{\ell}+n_u = n.
	\end{equation}
	Without loss generality, by removing small constants, therefore, the functional is defined as
	\begin{equation}
		R(A) = (K_\mathbf{F} K A)^\top (K_\mathbf{F} K A) - (K_\mathbf{F} K A)^\top K_\mathbf{D} Y + \lambda A^\top K A + Y^\top K_\mathbf{D}^\top K_\mathbf{D} Y,
	\end{equation}
	with constraints
	\begin{equation*}
		0 \le KA \le 1,
	\end{equation*}
	The $\lambda$ is the regularization parameter, $K_\mathbf{F}=K(x_i,x_j)$ is the positive semi-definite Fredholm kernel for $1 \leq i \leq n_\ell$,$1 \leq j \leq n$. $K=K(x_i,x_j)$ is the symmetric target kernel and $K_\mathbf{D}=K(x_i,x_j)$ represents the associated kernel for the data function for $1 \leq i,j \leq n$.
	However, the closed-form solution for the above optimization problem can be obtained as
	\begin{equation}
		A = ( K_\mathbf{F}^\top K_\mathbf{F} K + \lambda I)^{-1} K_\mathbf{F}^\top K_\mathbf{D} Y
	\end{equation}
	The "noise assumption" in semi-supervised learning refers to the idea that the neighborhood of every point in the data should contain information about the conditional distribution of the labels, but that the directions\footnote{refers to the axes or features of the data points in the feature space.} with lower variances in the unlabeled data can be regarded as noise.The direction with lower variance in the unlabeled data is considered to be lacking information with respect to the class labels because it is believed to contain only random variations and not any useful information. The low variance in this direction may be a result of the absence of significant changes in the data or the presence of irrelevant features. In either case, these directions are considered to be uninformative or "noisy" with respect to the class labels, which can negatively impact the performance of a supervised learning algorithm.\\
	Our primary objective is not to examine the impact of semi-supervised learning assumptions on the problem, as this information is readily available in the references. Rather, we aim to demonstrate that utilizing distribution dependent operators without considering the impact of the probability distribution may be impractical. Our goal here is to illustrate that varying kernels for the data function $G(T)$ offer a more effective means of controlling the high-dimensional Fredholm integral equation with solution (\ref{sol}), which is highly susceptible to over-fitting \cite{Belkin1}.

	\section{Experimental validation}
	\subsection{experimental settings}
	In this study, the proposed methods are evaluated through a comprehensive comparison with several established techniques, including Kernel Regularized Least Square (KRLS), $\mathcal{V}$-matrices \cite{VapnikV}, Fredholm learning \cite{f1} and Manifold Regularization \cite{Belkin1}. To assess their performance, we employ 12 classification datasets in table (\ref{table1}) obtained from the University of California, Irvine (UCI) machine learning repository. The datasets were selected based on previous work \cite{semi1} and some were binarized to fit our needs. To ensure fairness in comparison, the datasets underwent z-score normalization, with zero mean and a standard deviation of one. Additionally, the test set was normalized based on the mean and standard deviation of the training set. The Gaussian radial basis function (RBF) is used as a kernel function in the similarity computation, and the RBF width factor ($\sigma$) was set equal to the dimensionality of the datasets, which is a technically appropriate scaling. In an effort to determine the optimal regularization parameter and related parameters in the MSDF method, we utilized the "expand.grid" approach to capture all possible parameters for the eight methods. This was followed by a 10x5-fold cross-validation, which was performed to experimentally validate the results.
	\begin{table}[h!]
		\setlength{\tabcolsep}{20pt}
		\centering
		\begin{tabular}{llll}
			\hline
			Dataset & Name & Size & Attr \\ [0.5ex] 
			\hline
			1 & Abalone & 4177 & 9 \\ 
			2 & Blood & 748 & 5 \\
			3 & Breast cancer & 699 & 11\\
			4 & Bupaliver & 345 & 7 \\
			5 & Cmc & 1473 & 10  \\
			6 & Heart & 270 & 14 \\
			7 & Pima & 768 & 9 \\
			8 & Sonar & 208 & 61  \\
			9 & Statlog & 690 & 15  \\
			10 & Survival  & 306 & 4 \\
			11 & Vehicle & 846 & 19 \\
			12 & Wdbc & 569 & 32 \\ [1ex] 
			\hline
		\end{tabular}
		\caption{12 classification datasets taken from the University of California, Irvine (UCI) machine learning repository}\label{table1}
	\end{table}
	The regularization parameter $\lambda$ was varied between $1\times 10^{-4}$ and $1\times 10^{2}$, and an appropriate value was selected for the semi-supervised Laplacian and other MSDF kernels. In the case of the MSDF method, we implemented four different convolution kernels based on Table (\ref{table2}).
	\begin{table}[h!]
		\setlength{\tabcolsep}{20pt}
		\centering
		\begin{tabular}{ll}
			\hline
			Kernel & $K(x,x')$ \\ [0.5ex] 
			\hline
			The Gaussian RBF & $M_1 = exp \big(\sigma ||x-x'||^2 \big)$ \\ 
			The Laplacian & $M_2 =exp \big(\sigma ||x-x'|| \big)$ \\
			The Bessel & $M_3 =Bessel_{(\nu+1)}^n \sigma ||x-x'||$ \\
			The Anova RBF & $M_4 = \sum_{i \le i_1 < i_D \le N} \prod_{d=1}^{D} k(x_{id},x'_{id})^{*}$ \\ [1ex] 
			\hline
		\end{tabular}
		\caption{In the implementation of the MSDF method, four convolution kernels were selected to determine the operator kernel and data function in an arbitrary manner. The notation $M_{ij}$ indicates that the machine was constructed with operator kernel $M_i$ and data function $M_j$. $k(x_{id},x'_{id})$ represents a Gaussian kernel*.}\label{table2}
	\end{table}
	Table (\ref{table3}) shows that we used expand.grid to identify the best possible combinations of kernels for the operator and data functions, as well as their corresponding parameters for a specific dataset. It should be noted that the complexity of $\mathcal{V}$-Matrices is $\mathcal{O}(m^2 d^2)$ and that of $\mathcal{SV}$-Matrices is $\mathcal{O}(m^2 nd^2)$, which incurs significant computational costs. Therefore, in order to calculate these matrices, we used c++ environment for time consuming (for more details and accessing the $\mathcal{SV}$-Matrices visit \hypersetup{
		colorlinks=true,
		linkcolor=blue,
		urlcolor=blue,
	}
	\url{https://github.com/AxiomMachine/V-Matrix-functions}).\\
	The experiments in this study were divided into three parts. In the first part, we evaluated the classification performance of all twelve datasets using the considered methods. However, the models were trained with only a few training samples in order to monitor their performance and evaluate the ability of the models to handle limited datasets. As a result, we observed that some models were unable to generalize and failed for different datasets.\\
	In the second part of the experiments, we selected four datasets from Table (\ref{table1}) and varied the regularization parameter to create high-variance and low-variance (high-bias) models. This was done to investigate the stability of the models under different levels of regularization.
	\begin{table}[h!]
		\begin{tabular}{p{15cm}}
			\hspace{15em} \textbf{The MSDF Method Algorithm} \\
			\hline
			\textbf{Input}: training data $(x_{1}, y_{1}), \ldots ,(x_{n_\ell}, y_{n_\ell})$, unlabeled observations ${x_{u}}$.\\
			\textbf{Output}: Accuracy, nominated form of kernel and its parameters.\\
			\hline
			\textbf{Algorithm}: \\
			\hspace{1em} 1. Initialize a list of 16 different combinations of kernels \\
			\hspace{1em} 2. Use \texttt{expand.grid} to incorporate all combinations of parameters \\
			\hspace{1em} 3. \textbf{For} each fold in the cross-validation \\
			\hspace{2em} a. \textbf{For} each combination of parameters of the kernels\\
			\hspace{3em} i. \textbf{For} each combination of kernels and their parameters\\
			\hspace{4em} 1. Fit the model\\
			\hspace{4em} 2. Calculate the accuracy of the model \\
			\hspace{2em} c. Store the accuracy and the nominated form of the kernels and their parameters \\
			\hspace{1em} 3. Select the model with the lowest accuracy \\
			\hspace{1em} 4. Return accuracy and nominated form of kernel and its parameters \\
			\hline
		\end{tabular}
		\caption{The MSDF method algorithms show that all combinations of kernels and corresponding parameters should be investigated comprehensively.}\label{table3}
	\end{table}
	In the last part of the experiment, we selected the "Sonar" dataset, which has 61 dimensions and 208 samples and is considered complex enough to test the competitiveness of the models. We trained the eight methods with appropriate regularization and the MSDF method kernels parameters. The experiment was conducted using different training sizes (1\%, 25\%, 50\%, and 75\%) to evaluate how well the models can generalize with varying sample sizes.

	\subsection{Experimental results}
\begin{figure}
		\centering
		\begin{minipage}[b]{0.4\textwidth}
			\centering
			\hspace{-2cm}\Large\textbf{High-Variance}
			\includegraphics[width=\linewidth]{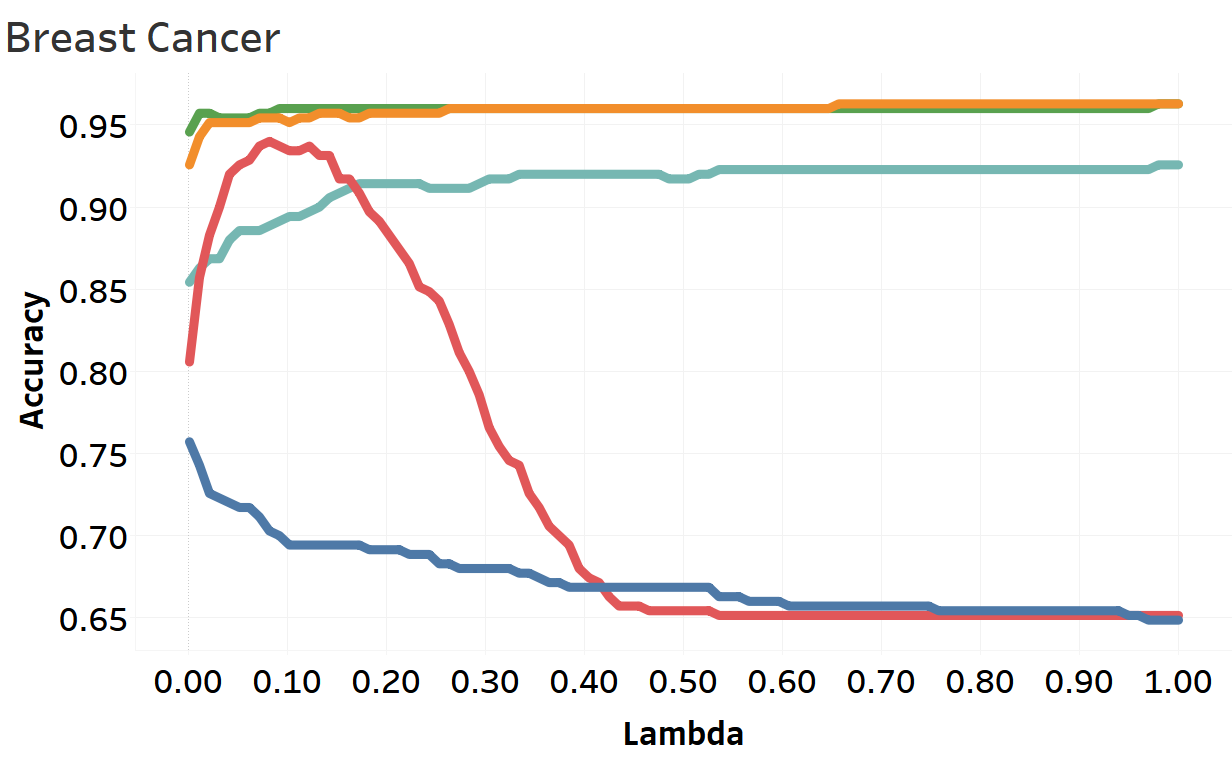}\\
			\includegraphics[width=\linewidth]{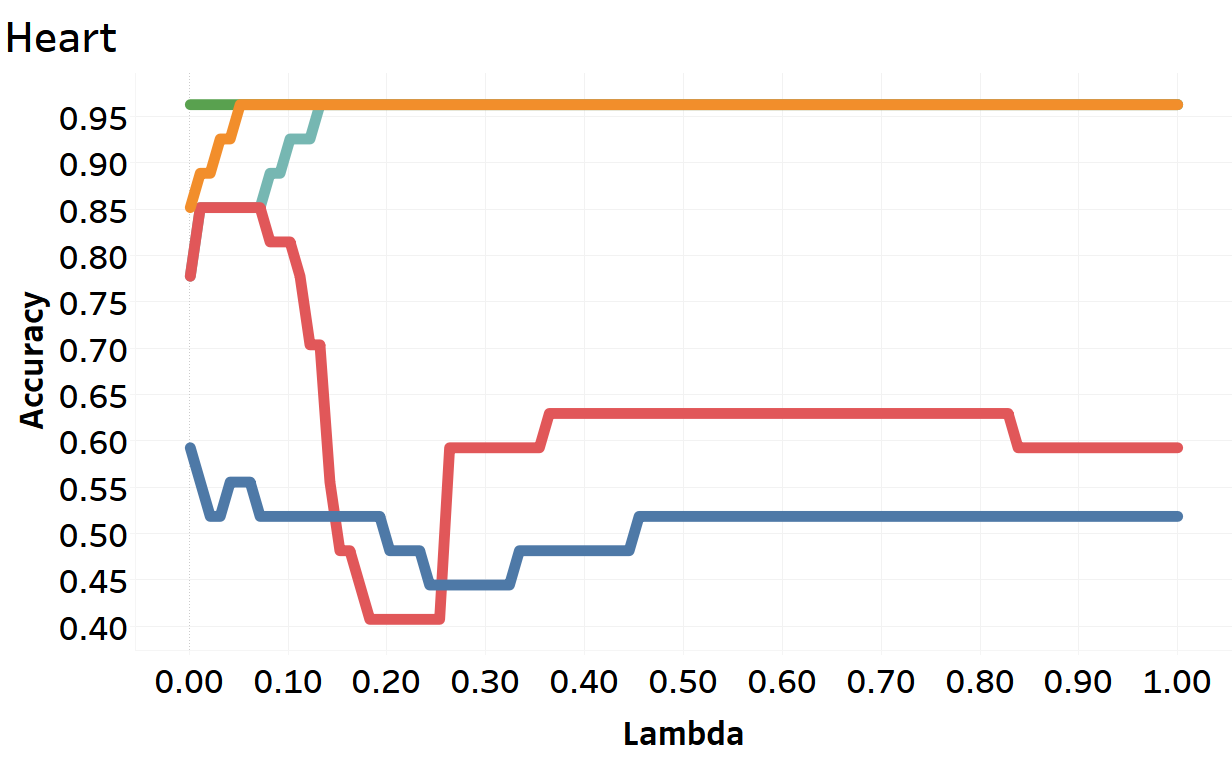}\\
			\includegraphics[width=\linewidth]{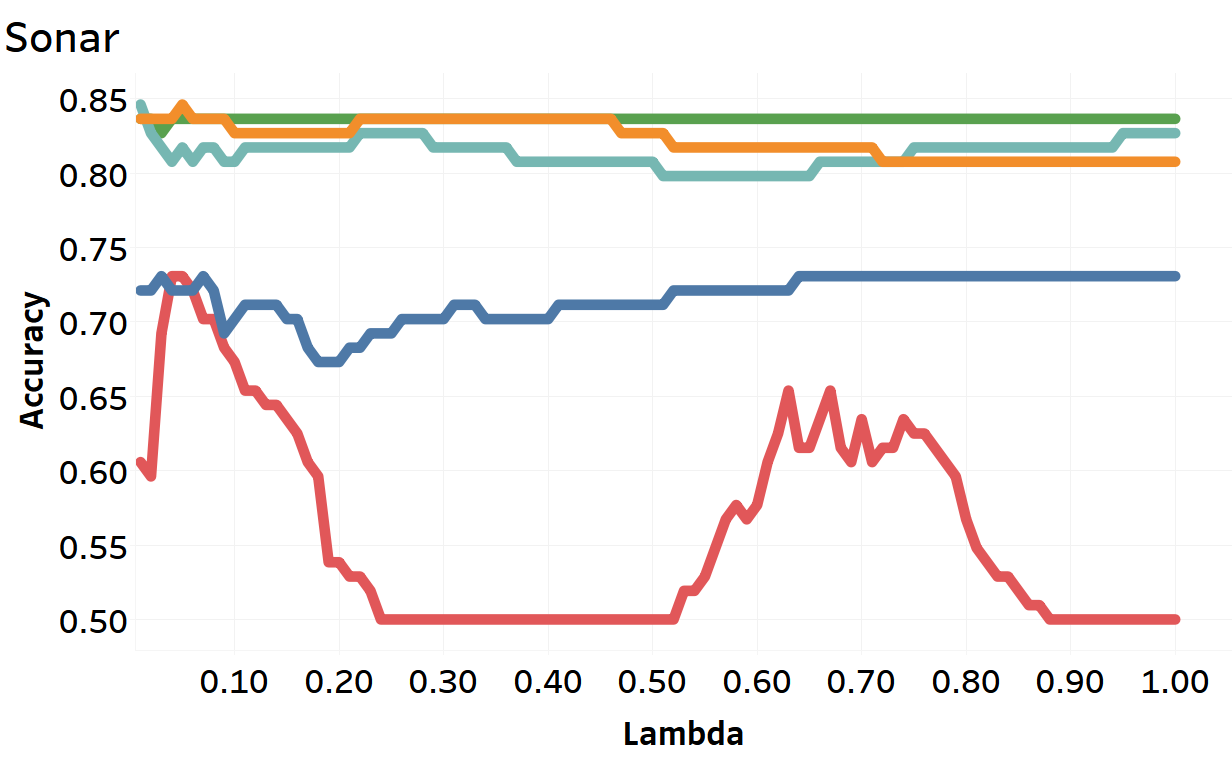}\\
			\includegraphics[width=\linewidth]{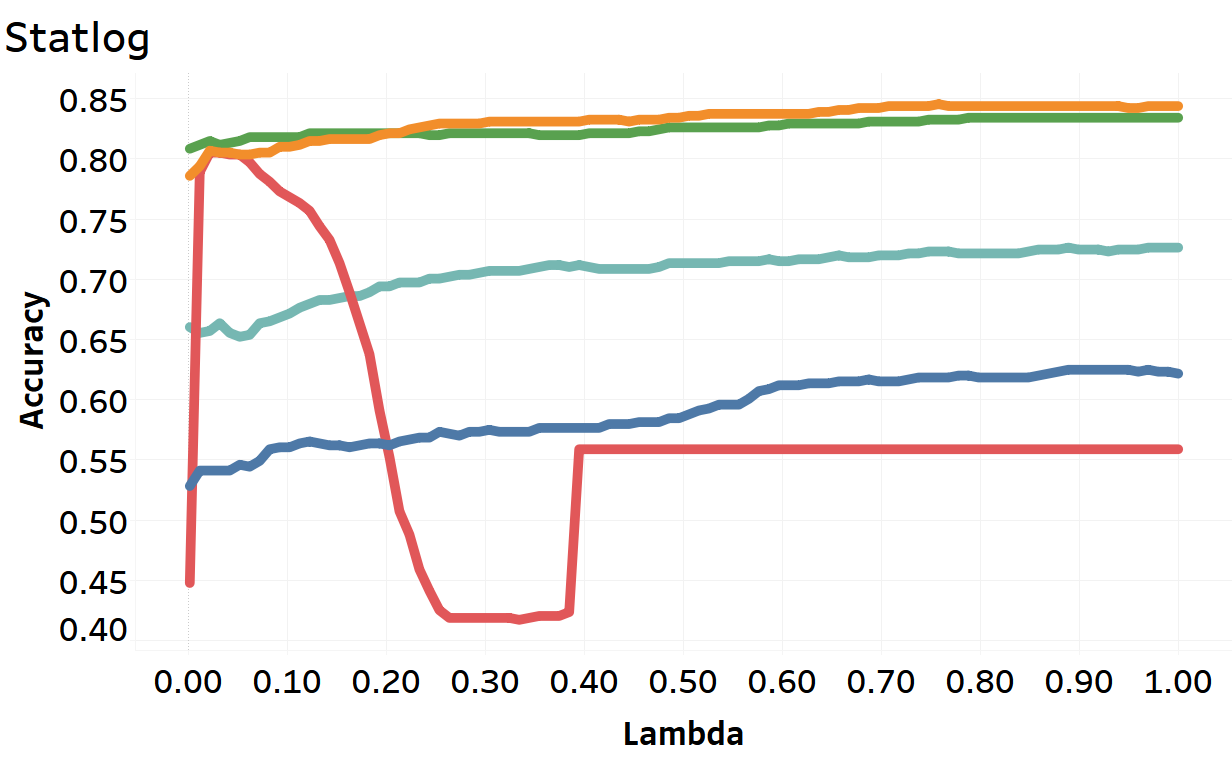}
		\end{minipage}%
		\begin{minipage}[b]{0.4\textwidth}
			\raisebox{1.5\baselineskip}{\Large\textbf{High-Bias}}
			\includegraphics[width=\linewidth]{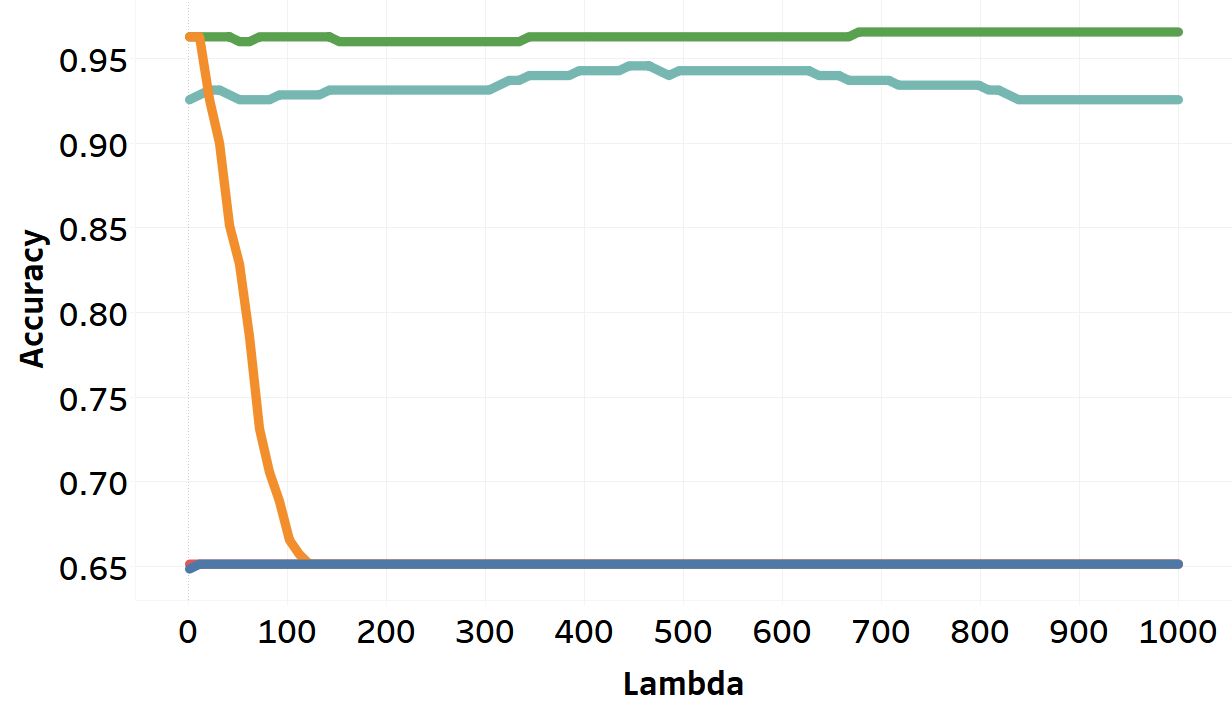}\\
			\includegraphics[width=\linewidth]{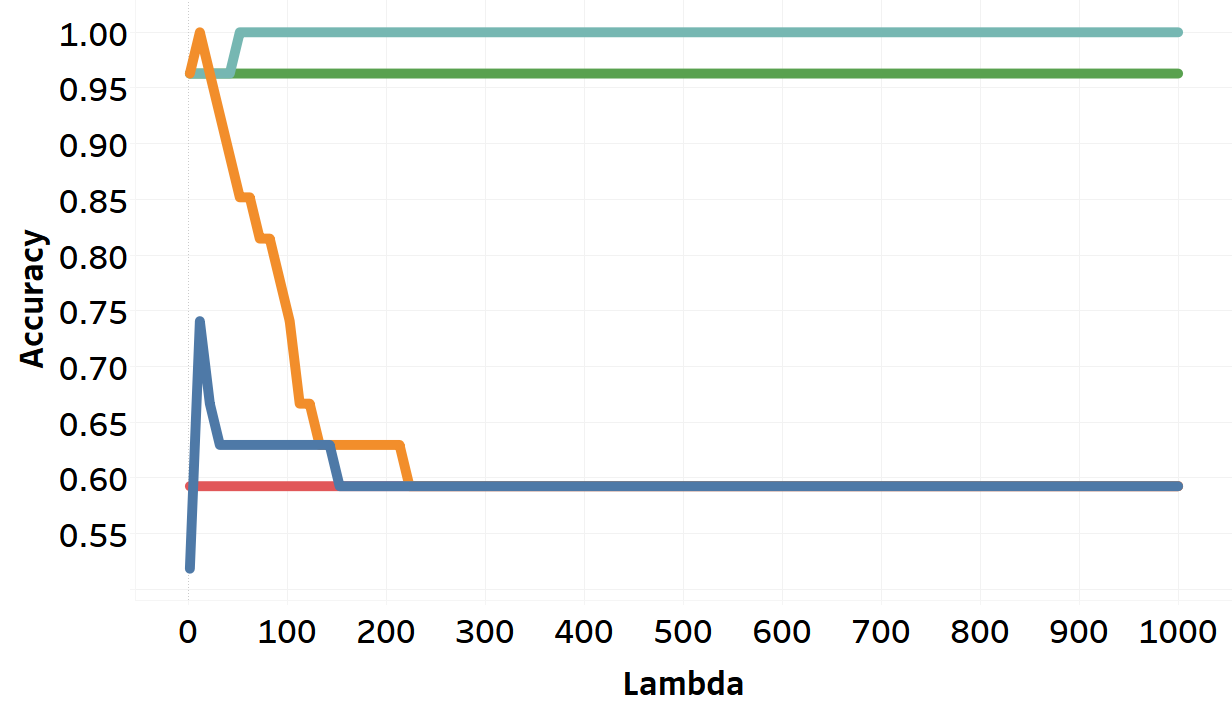}\\
			\includegraphics[width=\linewidth]{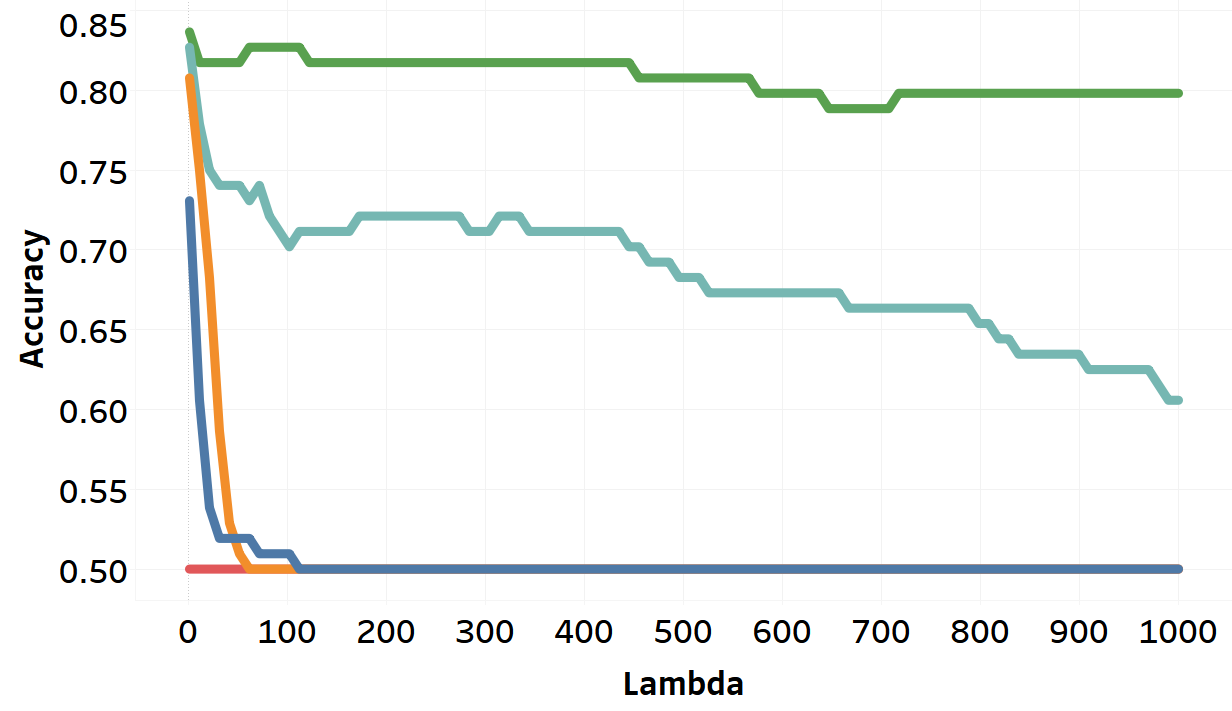}\\
			\includegraphics[width=\linewidth]{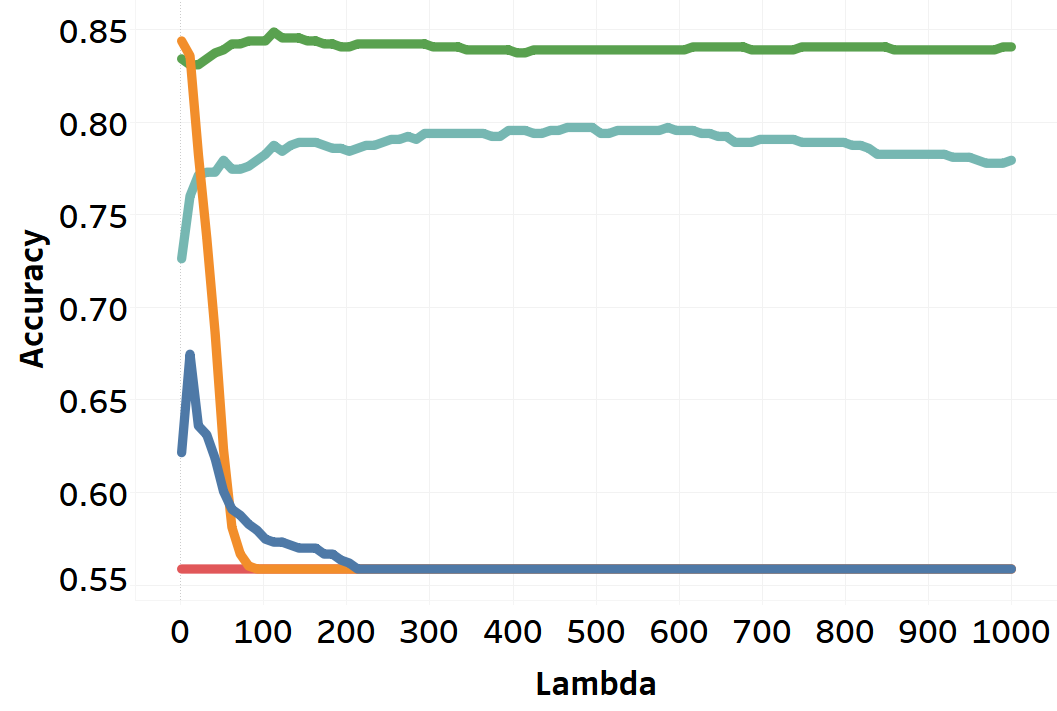}
		\end{minipage}
		\includegraphics[width=10cm]{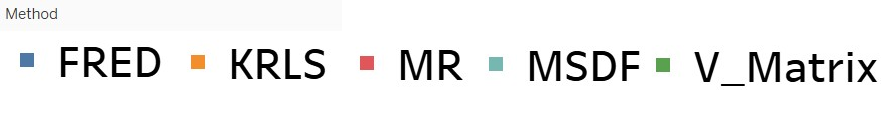}
		\caption{We evaluated the performance of models under complexity using the Breast Cancer, Heart, Sonar, and Statlog datasets. The figure displays the results for high-bias models (right panel) and high-variance models (left panel) on the four datasets.}
		\label{fig3}
	\end{figure}

\begin{figure}[h]
		\includegraphics[width=13cm]{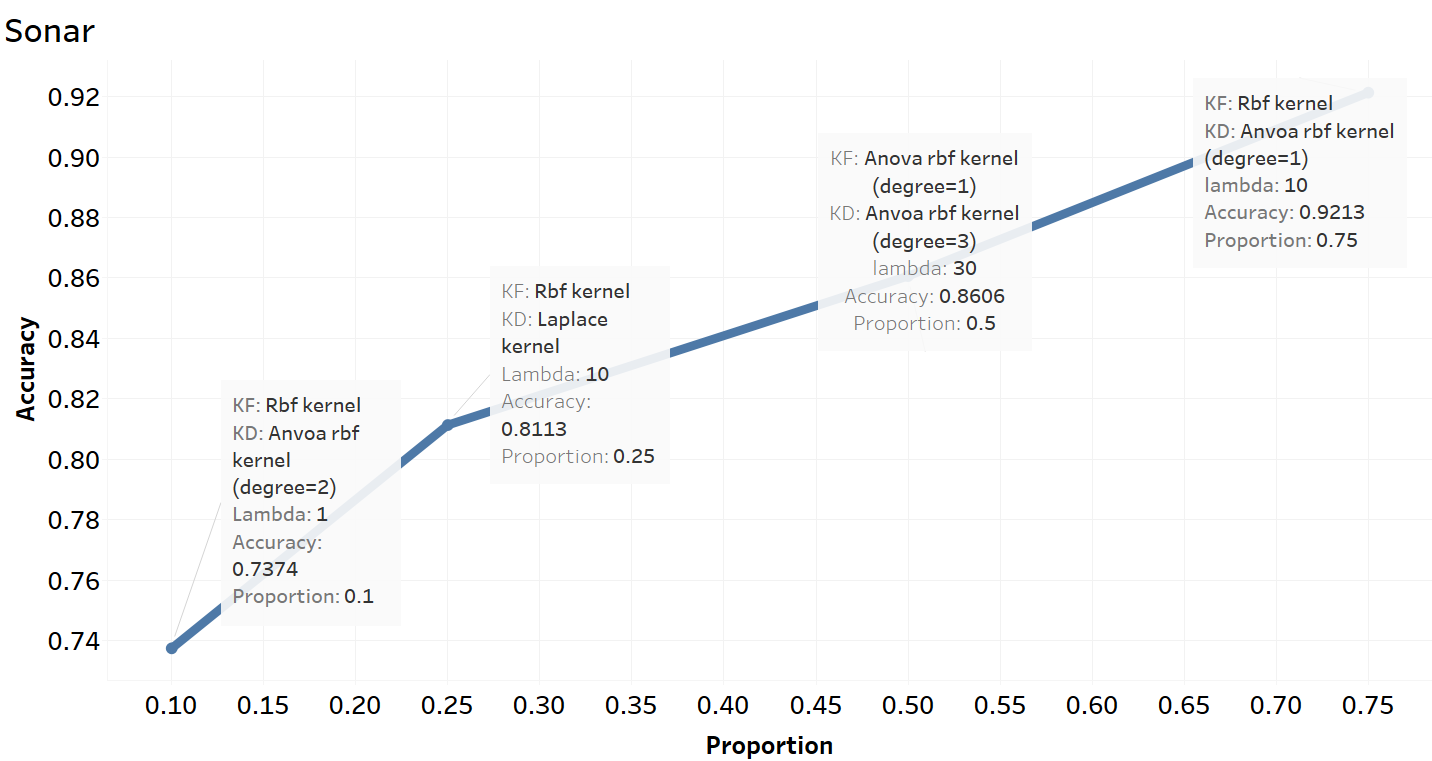}
		\includegraphics[width=13cm]{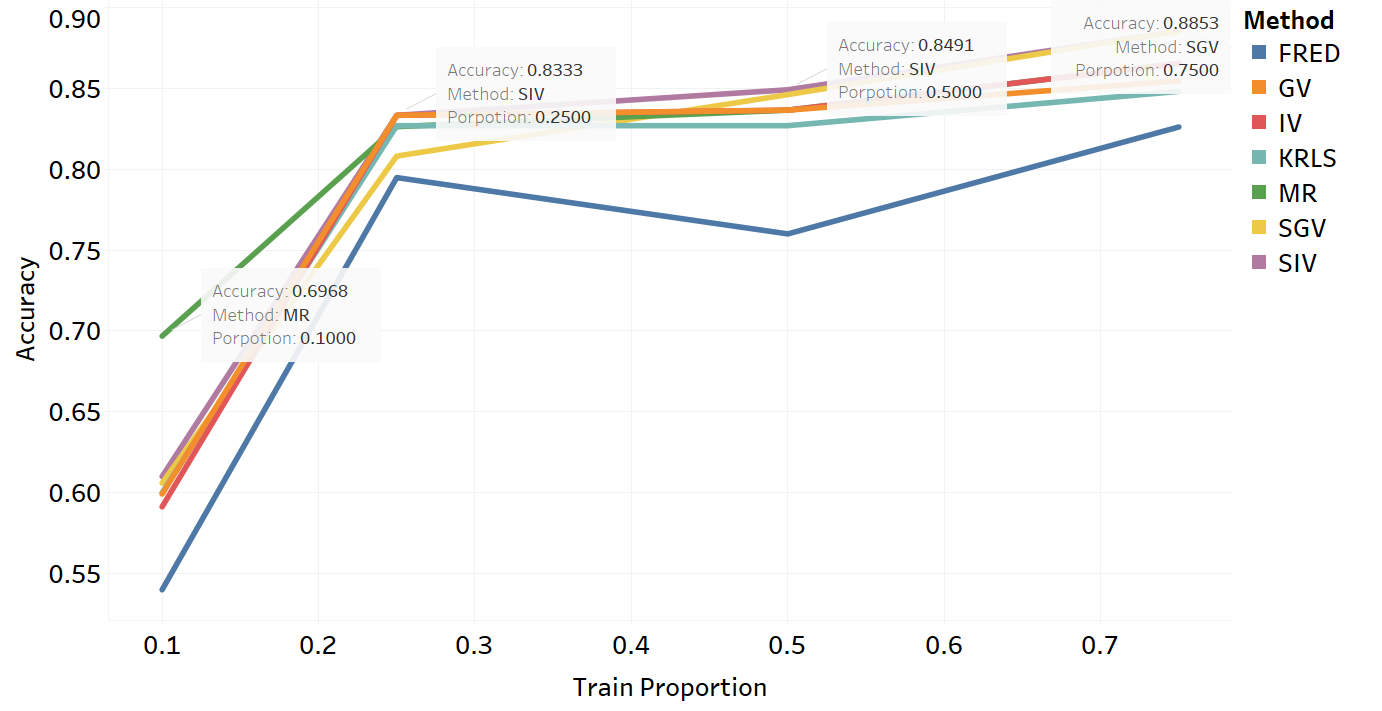}
		\centering
		\caption{The first plot displays the classification performance of the MSDF method using different sample proportions, while the plot below presents the performance of other methods. The figure provides information such as accuracy, sample proportion, and the maximum performance achieved by the models.}
		\label{fig4}
	\end{figure}

	In Table (\ref{table4}), we present the average AUC of the eight compared methods that were tested on each of the twelve UCI datasets using 10$\times$5-fold CV. It should be noted that all free parameters, such as manifold parameter, regularization, and kernel parameters as shown in Table (\ref{table1}), were selected through cross-validation.\\
	The results indicate that in almost every dataset, there is an $M_{ij}$ method that exhibits significantly higher generalization performance compared to the other methods. However, the $\mathcal{SV}$-Matrices method demonstrated relatively better performance than KRLS, MR, and $\cal V$-Matrices. Nevertheless, the MDF method outperforms all the other methods by a significant margin, since out of the 16 arbitrary learning machine frameworks that were considered, at least one achieved high accuracy. It is noteworthy that the number of training data was minimized across all datasets.
	\begin{table}[ht]
		\centering %
		\begin{tabular}{c c c c c c c c c} %
			\hline\hline %
			Dataset & KRLS & SIV & SGV & MSDF & MR & FRED & IV & GV \\ [0.5ex] %
			\hline\hline %
			1 & 0.47 & 0.54 & 0.54 & \textbf{0.61} ($M_{12}$) & 0.56 & 0.58 & 0.53 & 0.54 \\ %
			2 & 0.46 & 0.49 & 0.48 & \textbf{0.58} ($M_{33}$) & 0.54 & 0.54 & 0.54 & 0.49 \\
			3 & 0.84 & 0.82 & 0.84 & \textbf{0.90} ($M_{42}$) & 0.65 & 0.80 & 0.78 & 0.84 \\
			4 & 0.48 & 0.55 & 0.56 & \textbf{0.74} ($M_{12}$) & 0.55 & 0.58 & 0.56 & 0.59 \\
			5 & 0.43 & 0.52 & 0.53 & \textbf{0.74} ($M_{41}$) & 0.49 & 0.53 & 0.54 & 0.53 \\
			6 & 0.76 & 0.78 & 0.78 & \textbf{0.83} ($M_{41}$) & 0.62 & 0.78 & 0.75 & 0.78 \\ %
			7 & 0.54 & 0.61 & 0.60 & \textbf{0.63} ($M_{32}$) & 0.57 & 0.62 & 0.61 & 0.60 \\
			8 & 0.51 & 0.72 & 0.71 & \textbf{0.74} ($M_{14}$) & 0.49 & 0.67 & 0.73 & 0.71 \\
			9 & 0.72 & 0.77 & 0.77 & 0.80 ($M_{42}$) & 0.63 & \textbf{0.81} & 0.70 & 0.71 \\
			10 & 0.55 & 0.60 & 0.60 & \textbf{0.77} ($M_{32}$) & 0.57 & 0.57 & 0.53 & 0.59 \\
			11 & 0.75 & 0.69 & 0.71 & 0.70 ($M_{12}$) & \textbf{0.80} & 0.76 & 0.74 & 0.71 \\
			12 & 0.88 & 0.85 & 0.87 & \textbf{0.89} ($M_{42}$) & 0.67 & 0.88 & 0.86 & 0.86 \\ [1ex] %
			\hline %
		\end{tabular}
		\caption{12 dataset binary classification are investigated for eight different methods. The accuracy of MSDF method is calculated correspond to the best $M_{ij}$ among 14 different methods according to the table (\ref{table1}). The index $i$ represents the operator kernel and the index $j$ represents data kernel.}\label{table4}
	\end{table}
	Figure (\ref{fig3}) illustrates the high stability of the $\mathcal{V}$-Matrix method, which can be attributed to the mutual residual in Eq (\ref{eq21}). Notably, the form of the $\mathcal{V}$-Matrix does not significantly affect its performance; rather, its stability can be attributed to the structure of the risk functional used in this method. In contrast, Fredholm learning and manifold regularization methods exhibit high instability with changes in regularization, indicating that their risk functionals are not appropriate for semi-supervised models (\ref{sol}). As a result, the stability of these methods is extremely fragile. While the MSDF method exhibits good performance, the $\mathcal{V}$-Matrix method maintains a nearly constant level of accuracy throughout this experiment.\\
	In the last part of these experiments, the figure (\ref{fig4}) consists of two plots. The first plot shows the MSDF method performance across four different levels of training sample sizes (1\%, 25\%, 50\%, and 75\%). Through this analysis, we found that certain combinations of operator and data function kernels result in higher accuracy compared to the other methods. For instance, when training with only 1\% of the data, the Fredholm kernel with an RBF data function kernel with degree=2 yielded the best results. Notably, the choice of kernels varied across different proportions of training data. Overall, the MSDF method consistently demonstrated better generalization performance than the other seven methods across all levels of training sample sizes.
	\section{Conclusion}
	When working with limited datasets, it is crucial to incorporate statistical inference into the supervised machine learning process in order to carefully design the problem. We demonstrated that defining an appropriate inverse mapping is an essential intermediate step in constructing a supervised learning problem for a small dataset. On the left-hand side of this linear operator, we defined the distribution-dependent Fredholm integral using an arbitrary kernel function, while the right-hand side defined the data distribution with an arbitrary kernel to approximate the exact datum. Indeed, the right-hand side of the Fredholm equation provides prior information that can regularize an ill-posed problem by ensuring the estimability of the solution, rather than forcing it into a specific interval. We demonstrated that the prior information $G$ can be selected arbitrarily to suit the desired problem. In this study, we aimed to incorporate the stochastic nature of the input or data function when estimating a conditional probability.\\
	To ensure the estimability and identifiability of the solution in a general framework, we addressed the stochastic ill-posed nature of the problem. However, without incorporating semi-supervised assumptions, the interfacing of data distributions captured by the Fredholm integral equation would be redundant.\\
	In the MSDF method, we selected four convolutional kernels, resulting in 16 distinct methods for estimating the desired solution. We selected the parameters of these kernels via cross-validation on a real dataset, leveraging prior information about the data distribution to select kernel parameters that compose a desired closed-form solution..
	\\\\\\\\

	\bibliography{references}

\end{document}